# REAL-TIME PEDESTRIAN DETECTION APPROACH WITH AN EFFICIENT DATA COMMUNICATION BANDWIDTH STRATEGY

**Mizanur Rahman*, Ph.D.**
Postdoctoral Fellow
Center for Connected Multimodal Mobility ($C^2M^2$)
Glenn Department of Civil Engineering, Clemson University
127 Lowry Hall, Clemson, SC 29634
Tel: (864) 650-2926; Email: mdr@clemson.edu

**Mhafuzul Islam**
**Ph.D. Student**
Glenn Department of Civil Engineering, Clemson University
351 Flour Daniel, Clemson, SC 29634
Tel: (864) 986-5446; Fax: (864) 656-2670;
Email: mdmhafi@clemson.edu

**Jon Calhoun, Ph.D.**
**Assistant Professor**
Clemson University
Holcombe Department of Electrical and Computer Engineering
221-C Riggs Hall, Clemson, SC 29634
Tel: (864) 656-2646, Fax: (864) 656-5910
Email: jonccal@clemson.edu

**Mashrur Chowdhury, Ph.D., P.E., F.ASCE**
**Eugene Douglas Mays Endowed Professor of Transportation**
Clemson University
Glenn Department of Civil Engineering
216 Lowry Hall, Clemson, South Carolina 29634
Tel: (864) 656-3313   Fax: (864) 656-2670
Email: mac@clemson.edu

*Corresponding author

**ABSTRACT**
Vehicle-to-Pedestrian (V2P) communication can significantly improve pedestrian safety at a signalized intersection. It is unlikely that pedestrians will carry a low latency communication enabled device and activate a pedestrian safety application in their hand-held device all the time. Because of this limitation, multiple traffic cameras at the signalized intersection can be used to accurately detect and locate pedestrians using deep learning and broadcast safety alerts related to pedestrians to warn connected and automated vehicles around a signalized intersection. However, unavailability of high-performance computing infrastructure at the roadside and limited network bandwidth between traffic cameras and the computing infrastructure limits the ability of real-time data streaming and processing for pedestrian detection. In this paper, we develop an edge computing based real-time pedestrian detection strategy combining pedestrian detection algorithm using deep learning and an efficient data communication approach to reduce bandwidth requirements while maintaining a high object detection accuracy. We utilize a lossy compression technique on traffic camera data to determine the tradeoff between the reduction of the communication bandwidth requirements and a defined object detection accuracy. The performance of the pedestrian-detection strategy is measured in terms of pedestrian classification accuracy with varying peak signal-to-noise ratios. The analyses reveal that we detect pedestrians by maintaining a defined detection accuracy with a peak signal-to-noise ratio (PSNR) 43 dB while reducing the communication bandwidth from 9.82 Mbits/sec to 0.31 Mbits/sec, a 31x reduction.

**INTRODUCTION**

According to National Highway Traffic Safety Administration (NHTSA), pedestrian fatalities are increasing in the United States every year. Pedestrians account for 14% of US road fatalities with over 5,376 annual fatalities in 2015 (based on the latest statistics related to pedestrian fatalities) *(1)*. Moreover, on average 69,000 pedestrians are injured annually on US roadways. Recent research work shows that enabling Dedicated Short Range Communication (DSRC), which is a low latency data communication medium for safety applications, in a pedestrian hand-held device can increase pedestrian safety significantly through vehicle-to-pedestrian (V2P) communications *(2)*. The DSRC-enabled V2P system gives a 360-degree view where both the driver and the pedestrian are warned of a possible collision using DSRC-based safety alerts. However, it is very unlikely that all the pedestrians carry a DSRC enabled device and activate a pedestrian safety application all the time. In addition, the current cellular communication network is not applicable for pedestrian safety applications because of high data exchange latency *(3)*.

Video and images are commonly used in traffic monitoring on a roadway and object detection applications in Intelligent Transportation Systems (ITS) *(4,5)*. Multiple traffic cameras at the signalized intersection can be used to accurately detect and locate pedestrians instead of the DSRC-enabled pedestrian hand-held device. Computing infrastructure can be used to process video data and detect pedestrians from the video data using the camera feed and then broadcast safety alerts related to pedestrians to warn vehicles around a signalized intersection. In a typical ITS deployment, where one or more transportation sensors (e.g., traffic cameras, roadway sensors) transmit sensing data (e.g., image, numerical, text sensor data) over a network to an ITS processing location (ITS center, such as a Traffic Management center) with substantially more processing capability. However, a centralized computing service cannot support real-time Connected Vehicle (CV) applications, such as pedestrian safety application, due to the often unpredictable network latency, high data loss rate, and expensive bandwidth, especially in a mobile environment like the CV environment *(4, 6)*.

Edge computing is a new computing concept that enables data analytics at the source of the data for real-time safety applications *(7, 8)*. Edge-based computing pushes the frontier of computing applications, data, and services away from centralized computing infrastructures to the edges. For example, a roadside data infrastructure located in the next immediate edge layer (e.g., roadside transportation infrastructure) from the associated CVs can offer computational ability to support CV safety applications.

However, network bandwidth between bandwidth-hungry roadside surveillance devices, especially from traffic cameras, and computing infrastructure limits the ability of real-time data streaming and processing for ITS applications. More efficient use of the bandwidth allows for the deployment of real-time vision-based object detection using deep learning algorithms. Understanding this constraint will enable us to build systems that will be widely deployed. Lossy data compression can significantly reduce data storage requirements for massive data volumes and decrease the data transmission time in a communication network of an ITS system. However, lossy compressed video data degrades the image quality and eventually reduces the chance of object detection, but does allow for a more efficient use of the bandwidth. Though currently, the computation ability of some devices is limited, as well as the communication, there is a chance they will be substantially improved in the near future. As the quality of communication and computation increases, lossy compression can still be an effective technique to reduce bandwidth requirements *(9)*. Reducing bandwidth requirements can allow for less expensive components being used lowing the price for the deployment *(10)*.

In this study, we developed an edge computing based real-time pedestrian detection strategy combining deep learning based pedestrian detection application and an efficient data communication approach to reduce bandwidth requirements while maintaining a high object detection accuracy. We use lossy video data compression techniques (i.e., Lossy Compression (LC)) for greater use of existing bandwidth or for greater resiliency with existing bandwidth for data transmission between video surveillance and computing infrastructure. Then we determine the tradeoff between the reduction of the communication bandwidth requirements and a defined object detection accuracy with traffic camera data collected from a signalized intersection in Clemson, SC, USA.

The remainder of this paper is structured as follows. Related studies section describes related work on vision-based object detection model and lossy compression for traffic camera data. Research method section presents the strategy for determining the compressibility of traffic camera data using a vision based pedestrian detection algorithm. Analysis of a lossy video compression based pedestrian detection strategy is presented in analysis and results section. Finally, last section provides a concluding discussion.

**RELATED STUDIES**

There is limited work evaluating the performance of lossy compression (LC) to problems in the ITS domain. The use of LC for video data requires a definition of correctness, and in the context of real-time object detection using vision-based techniques, the correctness metric is the classification precision and recall of the object detection model. Lossless compression reduces data size and incurs no information loss, but results in low compression ratios and compression/decompression bandwidths *(11)*. LC, however, significantly reduces the size of videos by introducing noise when representing each frame with fewer bits *(12)*. Error-bounded lossy compression allows for limiting the amount of loss via a user-defined error bound. The performance of lossy compression is typically faster than lossless compression and results in large compression ratios (i.e., smaller files) *(13)*. Because error-bounded lossy compression allows user control of the accuracy level, the tolerance can be dynamically modified to ensure the quality of service. The achievable compression bandwidth is therefore dependent on the magnitude of the LC's induced noise/error *(14)*. LC techniques for video data are commonly used in online streaming platforms such as Netflix and YouTube *(15)* but have yet to be investigated in the ITS domain. Because of its novelty in the ITS domain, there are several open questions about what is required from LC in the ITS domain with respect to the compression ratio, the compression bandwidth, acceptable amount of noise, and where in the ITS deployment. If LC is done, error accumulation that further degrades video quality is possible. Furthermore, the impact of error accumulation due to multiple LC operations remains unstudied.

There are many different video compression formats in common use: H.262, H.264, high efficiency video coding (HEVC), etc *(16)*. H.262 compresses each frame of a video by applying a discrete cosine transform to sub-blocks of the image and encoding the coefficients or using the previous and next frames and compressing the difference between the common sub-blocks *(17)*. H.264 compresses similarly to H.262 but uses a lower bit rate to achieve the same level of quality *(18)*. HEVC, like H.262 and H.264, identifies inter- and intra-frame regions of similarity, but is further optimized to lower the bitrate while keeping the same quality level. HEVC is intended for use with high-resolution videos: 1080p, 4K, and 8K resolution *(19)*. To create a video file, a video compression format is combined with an audio compression format into container format (video file) such as AVI, MP4, FLV. It is often useful to convert video files between various formats, or further improving the compression ratio for transmission or long-term storage. FFmpeg is a powerful cross-platform multimedia framework capable of decoding, encoding, transcoding, and filtering most audio and video formats *(20)*. In this study, we will use FFmpeg tool for investigating the tradeoff between lossy compression of video data and pedestrian detection accuracy using a deep learning model.

For traffic operational analysis, different types of algorithms, such as embedded algorithms for loop detector systems, computer vision-based algorithms and machine learning based algorithms, have been used for solving different traffic related problems *(21-24)*. Machine learning based algorithms improve the accuracy for traffic operational analysis compared to statistical methods, as it can learn from previous experience with similar roadway conditions. To detect an object, machine learning systems take a classifier for that object and evaluate it at various locations and scales in a test image. Systems like deformable parts models (DPM) use a sliding window approach where the classifier is run at evenly spaced locations over the entire image *(25)*. More recent approaches like Recurrent-Convolutional Neural Network (R-CNN) use region proposal methods to first generate potential bounding boxes in an image and then run a classifier on these proposed boxes. After classification, post-processing is used to refine the bounding boxes, eliminate duplicate detections, and rescore the boxes based on other objects in the scene *(26)*. These complex pipelines are slow and hard to optimize because each individual component must be trained separately. Recently, Redmon et al. developed an object detection model, You Only Look Once (YOLO), to detect objects in real-time *(27)*. This model can detect object with a single network evaluation unlike systems like R-CNN, which require thousands for a single image. Because of this mechanism, YOLO model is 1000 times faster than R-CNN and 100 time faster than Fast R-CNN. Thus, it is applicable for real-time pedestrian detection in a connected transportation system.

**RESEARCH METHOD**

Figure 1 presents the general framework for the real-time pedestrian detection using the YOLOv3 (YOLO Model - version 3) deep learning model combined with lossy data compression. We extract video data from roadside traffic monitoring cameras and use it as the input for the lossy data compression algorithm. After compression, we transfer the data to the edge-computing infrastructure and use a pre-trained and calibrated YOLOv3 model to detect pedestrians. From the YOLO model, we calculate how many pedestrians are detected every one tenth of a second. Using the field collected video data, we prepare ground truth data to evaluate the pedestrian detection accuracy. Then, using the YOLOv3 model output and ground truth data, we evaluate the pedestrian detection accuracy.

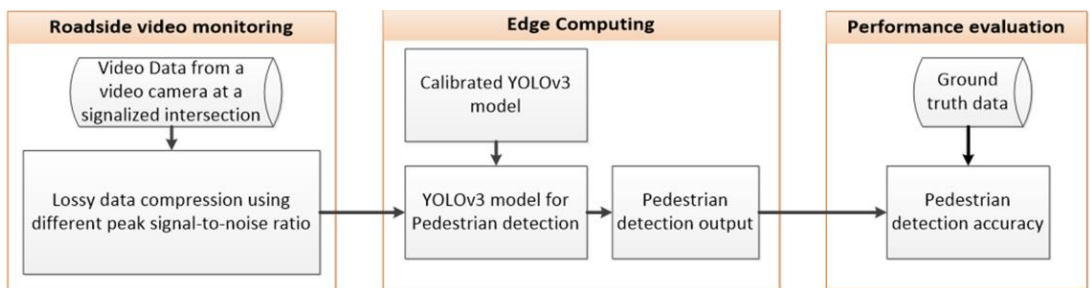

**FIGURE 1 Edge-computing based pedestrian detection strategy using a lossy data compression technique.**

**Edge Computing**

Edge computing is a new computing concept that enables data analytics at the source of the data for real-time safety applications (7, 8). Edge-based computing pushes the frontier of computing applications, data, and services away from centralized computing infrastructures to the edges. For example, a roadside data infrastructure located in the next immediate edge layer (e.g., roadside transportation infrastructure) from the associated CVs can offer computational ability to support CV safety applications. In general, an edge-centric connected vehicle systems consist of three edge layers (at least): i) mobile edge (e.g., connected vehicles); ii) fixed edge (e.g., roadside infrastructures); and iii) system edge (e.g., backend server at Traffic Management Center (TMC)) *(28)*.

CVs participating in our system will be acting as mobile edges, and it will be equipped with a DSRC-based On Board Unit (OBU). A fixed edge includes a data processing unit, such as an Intel® NUC device, which has a similar processing capability we have used in our experiments, and a Dedicated Short-Range Communication (DSRC) based roadside unit (RSU) that communicates with connected vehicles. A fixed edge can communicate with the mobile edges using DSRC and communicate with system edge using Optical Fiber/Wi-Fi. A system edge is a single end-point for a cluster of fixed edges. Fixed edge can be extended to support a video camera and other sensing devices, such as weather sensors and GPS sensors. Fixed edges are connected to a system edge that can effectively serve as a backend resource. In our connected vehicle environment, video camera installed at the intersection send video data to fixed edge (i.e., roadside infrastructure) and the pedestrian detection model will be implemented in a roadside unit (RSU) that includes a data processing unit, such as an Intel® NUC device, which can run the pedestrian detection model and a Dedicated Short-Range Communication (DSRC) based roadside unit (RSU) that communicates with connected vehicles (as shown in Figure 2).

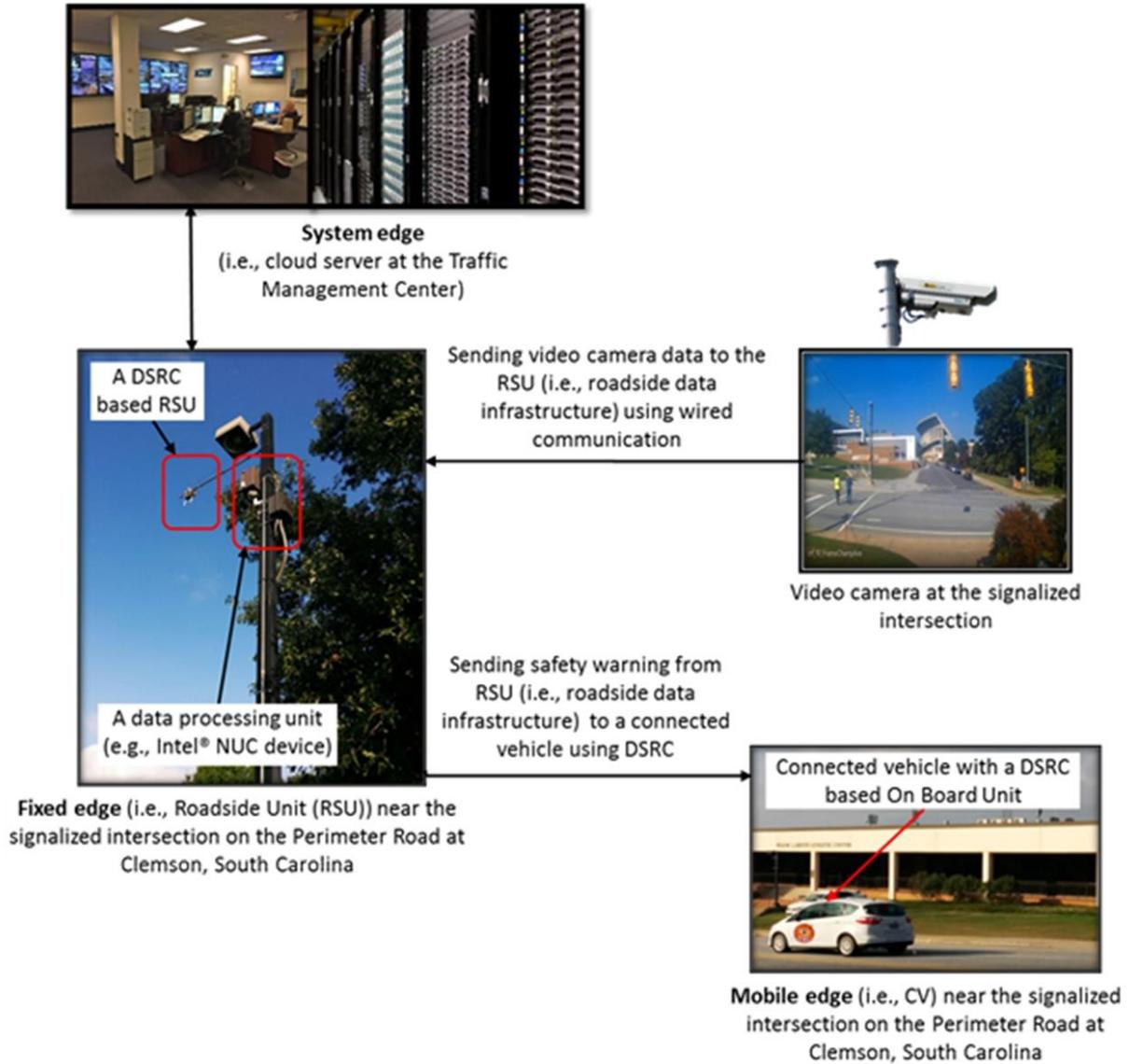

**FIGURE 2: Edge computing infrastructure for pedestrian detection.**

**Lossy Video Compression Strategy**

We utilize a commonly used lossy compression format H.264 on videos originating from a signalized intersection in Clemson, SC, USA to quantify the impact of lossy video compression in an ITS environment. To compress video data, we used the FFmpeg multimedia tool, and in particular, we use the Windows 64-bit binary release (N-82324-g872b358) *(29)*. FFmpeg is a versatile tool supporting operations on a wide variety of video formats and containers. FFmpeg provides an extensive command line interface for video transcoding, filtering, and streaming of video, images, and audio.

**YOLO Model for Pedestrian Detection**

*YOLOv3 model*
YOLOv3 model *(27)* divides an image into regions and predicts bounding boxes and probabilities for each region. These bounding boxes are weighted by the predicted probabilities. It looks at the whole image at test time so its predictions are informed by global context in the image. In addition, it performsobject detection with a single network evaluation unlike R-CNN, which requires thousands for a single image. The network architecture of the YOLO model was inspired by the GoogLeNet model for image classification *(24)*. The network has 24 convolutional layers followed by 2 fully connected layers. The YOLOv3 model uses 1×1 reduction layers followed by 3×3 convolutional layers *(30)*. Figure 3 presents the YOLOv3 model for pedestrian detection using field-collected video input. To process video images, this model resizes the input image to 448x448 and normalizes the image at the preprocessing layer. In this study, the size of our input image is 1280x720.

*YOLOv3 model calibration*
Initially, we use pre-trained YOLOv3 weights for pedestrian detection providing an 81% accuracy, which is not sufficient for a pedestrian safety application. To improve the pedestrian detection accuracy of the YOLOv3 model, we calibrate the model using the video data collected from traffic cameras. To train the YOLOv3 model, we extract 10 frames per second (fps) from the video. Next, we annotate each extracted frame to generate an annotated image file in the standard Pascal Visual Object Class (VOC) format *(31)*. In total, we collect 1300 images from 2 different intersections - (i) data collected for a signalized intersection with a small number of pedestrians and during sunny weather; and (ii) data collected for a signalized intersection with a large number of pedestrians and during cloudy weather – both located in Clemson, SC, USA. We use 900 images for training and 400 images for validation of the YOLOv3 model. However, after training the model, we find that it predicts multiple overlapping bounding boxes for a single pedestrian, which significantly reduces the pedestrian detection accuracy. To remove the overlapping bounding boxes and to improve the pedestrian detection accuracy, we implement a Non-max suppression method *(32)*. The Non-max suppression method takes a bounding box with a high confidence score and removes the overlapping regions using an Intersection of Union (IoU) value larger than 0.6. The IoU is calculated using following formulation:

$$IoU_{i,j} = \frac{O_{i,j}}{U_{i,j}}$$

where $IoU_{i,j}$ is the IoU value of bounding boxes $i$ and $j$, $O_{i,j}$ is the overlapped area of bounding boxes $i$ and $j$, and $U_{i,j}$ is the area of the union of bounding boxes $i$ and $j$. Using the re-trained YOLOv3 model and Non-max suppression method, the performance of pedestrian detection improves to 98% without any video compression.

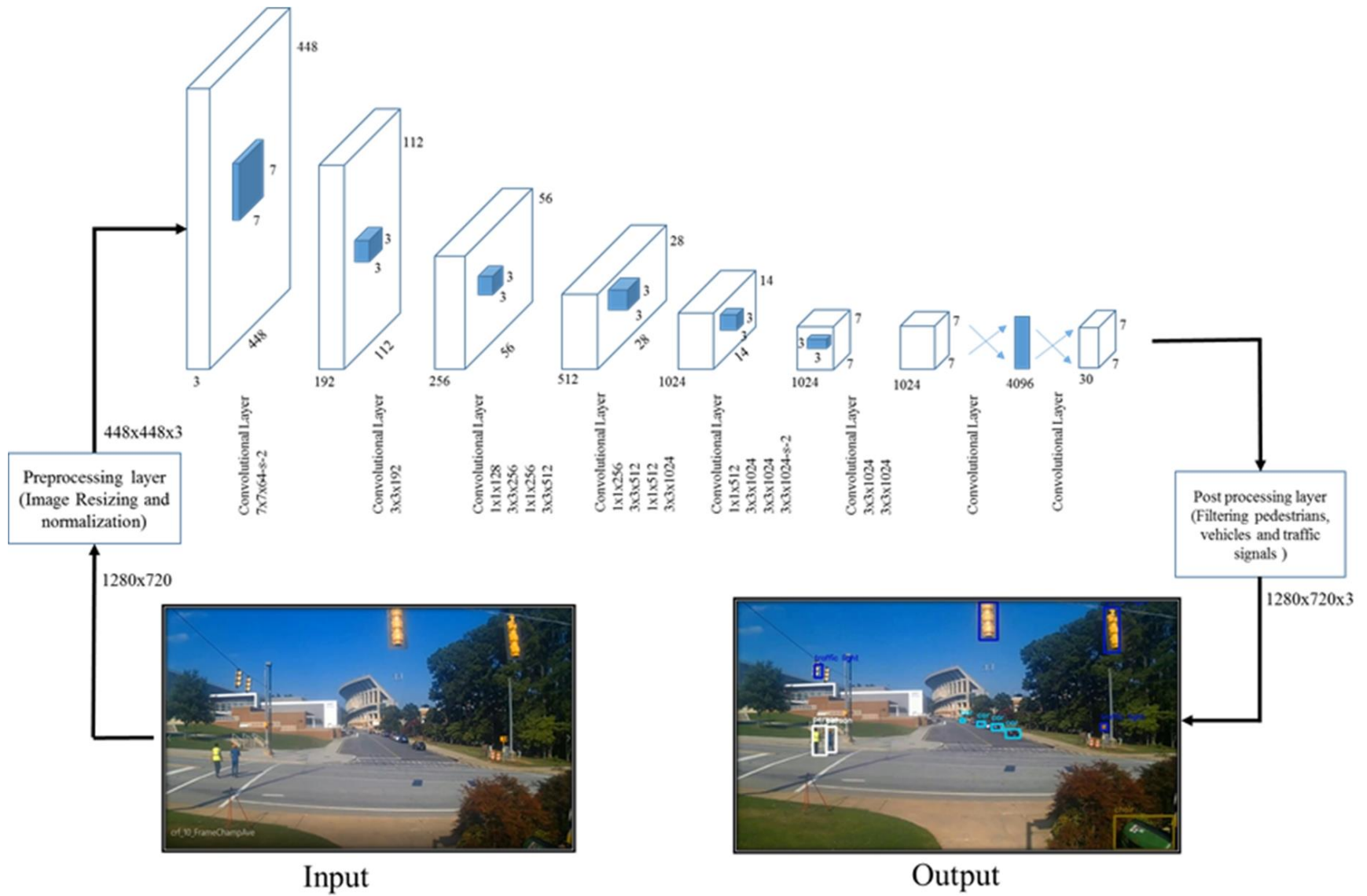

**FIGURE 3 Integration of YOLOv3 model with compressed video data for pedestrian detection.**

**Performance Measurements**

To evaluate the accuracy of our vision-based pedestrian detection strategy, we have measured the pedestrian classification accuracy for different video compression ratios. The accuracy indicates that the percentage of video frames where pedestrians are detected correctly. Manually labeled images are considered as the 'ground truth' value. For this binary classification problem, the classification accuracy is measured using the following formula for each test data set:

$$A = \frac{x}{n} * 100$$

where $A$ is the percentage of pedestrian detection accuracy, $x$ is the number of frames where pedestrians are correctly classified, and $n$ is the total number of frames for pedestrian detection events.

**ANALYSIS AND RESULTS**

**Data Description**

We have collected video data for two different scenarios: (1) Scenario 1 – data collected for a signalized intersection with a small number of pedestrians and during sunny weather; and (2) Scenario 2 – data collected for a signalized intersection with a large number of pedestrians and during cloudy weather. For scenario 1, we collected video data from the Perimeter Road and Avenue of Champions intersection at Clemson, SC, USA. It is a three-legged signalized intersection. We collect video data from three different directions. For each direction, we collected video data for 30 minutes from 5:30 PM to 6:00 PM. The size of each video data was 2 GB. For scenario 2, we collected video data from College Avenue and Highway 93 at Clemson, SC, USA. We collected data for one direction of roadway with a high number of pedestrians and the video was collected between 4:30 PM and 5:00 PM. For both scenarios, the raw videos from the cameras passes through a post-processing step that transforms the video from 30 frames per second (fps) to 10 fps (frame per second). In a connected vehicle environment, we collect Basic Safety Messages (BSMs) every one-tenth of a second to develop real-time safety-related applications. To mimic the standard for BSMs, we used 10fps for detecting pedestrians and broadcast safety alerts to the surrounding vehicles. To evaluate the performance of our detection model, we manually labeled each frame of the resulting videos to produce our ground truth data. The same 10 fps input video from the labeling step forms the input to the YOLO model. We used calibrated pre-trained weights for this model for pedestrian detection as described in the previous section.

**Lossy Video Compression Strategy**

Using field collected data, we have compressed video for different Constant Rate Factor (CRF) value to generate video data with different compression level. The range of the CRF value is 0 to 51; where 0 indicates no compression, and 51 is the highest compression level. After that we calculated Peak Signal-to-Noise Ratio (PSNR) by comparing original video file and compressed video file. PSNR is a well-known and used metric in digital signal processing. Thus, Constant Rate Factor (CRF) of FFmpeg is used to compress video yielding different compression ratios (i.e., small for CRF near 0 and large 51). However, it is important to know that the videos quality is measured by the PSNR and not CRF to make our results independent from the FFmpeg tool. Thus, it is required to determine at what PSNR value we are using for a specific CRF value. Table 1 provides a summary of average PSNR and CRF values for different evaluation scenarios.

**TABLE 1 Video Constant Rate Factor (CRF) and Peak Signal-to-Noise Ratio (PSNR)**

| Evaluation Scenarios | Constant Rate Factor (CRF) value for video compression | Average Peak signal-to-noise ratio (PSNR) |
|:---:|:---:|:---:|
| 1 | 10 | 56 dB |
| 2 | 20 | 49 dB |
| 3 | 30 | 43 dB |
| 4 | 33 | 41 dB |
| 5 | 35 | 40 dB |
| 6 | 37 | 39 dB |
| 7 | 40 | 37 dB |
| 8 | 50 | 31 dB |
| 9 | 51 | 30 dB |

Figure 4 presents compressed images from the field collected video data after FFmpeg compresses yielding different PSNR values. We observed that the quality of images is deteriorated with decreasing the PSNR values. Thus, as the video becomes more compressed, more noise is introduced that deteriorates quality. As the range of the CRF value is from 0 to 51, we varied the CRF value within that range and calculated a range of PSNR values from 30 dB to 56 dB. A decreasing PSNR makes identification of pedestrians more challenging and results in a lower probability of detection.

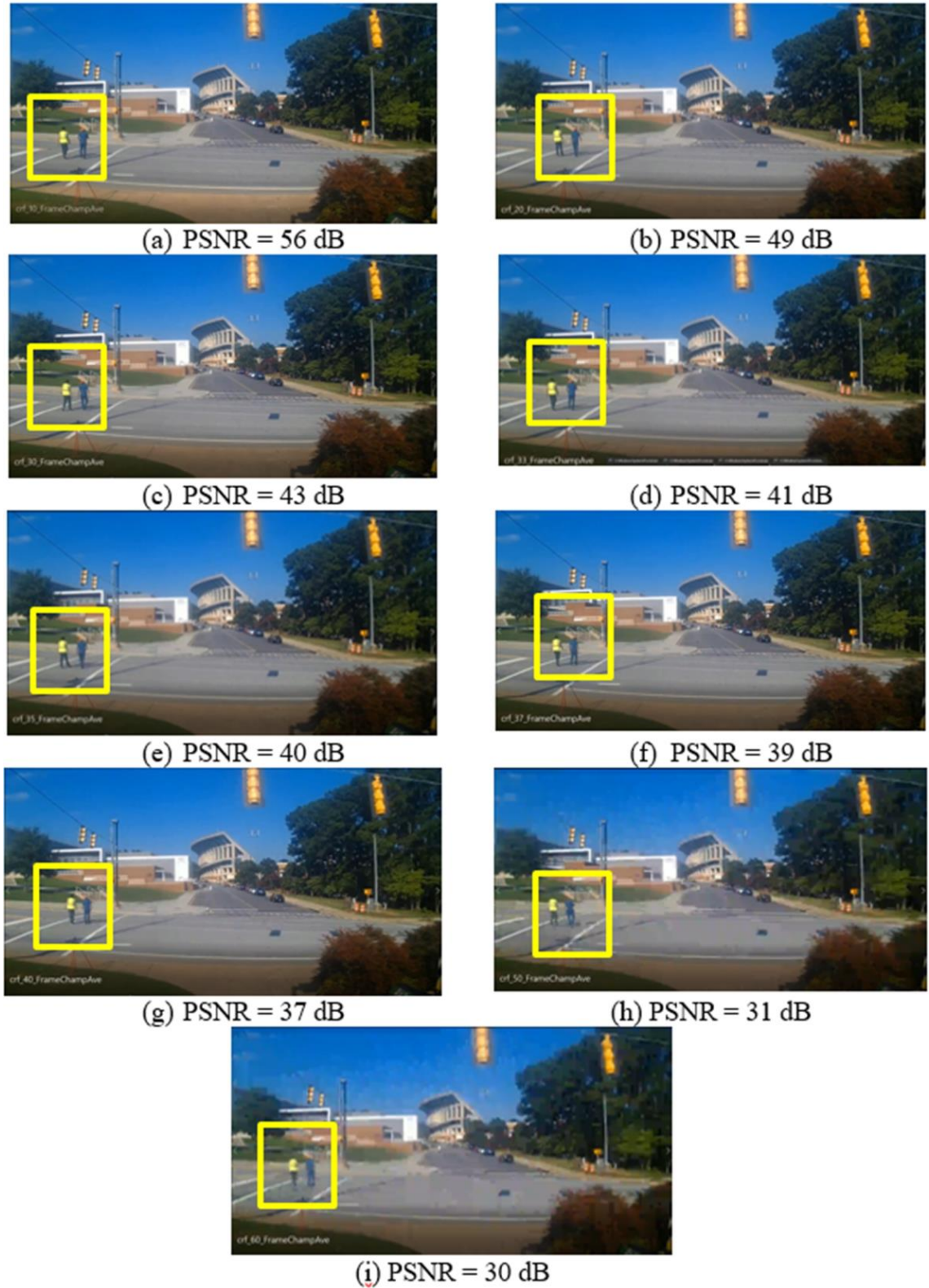

**FIGURE 4 Compressed images from the field collected data using increasing levels of video compression and different PSNR values.**

**Pedestrian Detection using YOLO Model**
Figure 5 presents a pedestrian detection output frame from the YOLO model using compressed field collected video data for scenario 1 with different PSNR values. As Figure 5 shows, we find that the YOLO model detects pedestrians accurately with PSNR value of 39 dB or higher. However, a comprehensive evaluation of the pedestrian detection model is required comparing with the ground truth data. For the comprehensive evaluation, we evaluate the pedestrian detection accuracy for different values of PSNR. Figure 6 shows that the maximum accuracy of pedestrian detection is 98% with the calibrated pre-trained weights and no compression for both scenario 1 and 2. For scenario 1, lossy compressed video with a PSNR of 56 dB, 49 dB and 43 dB, the detection accuracy remains constant at 98%. However, as the video becomes more distorted due to high levels of compression, the accuracy of the YOLO model starts decreasing. At the highest level of compression (PSNR 30 dB) the prediction accuracy is 60% for scenario 1 and 55% for scenario 2. This analysis shows that the threshold for compressing the data in terms of PSNR is 43 dB and it still maintains an acceptable pedestrian detection accuracy. Evaluation scenario 2 yields a similar pattern for pedestrian detection accuracy for different PSNR values. Therefore, the number of pedestrians does not affect the detection accuracy till PSNR value of 43 (as shown in Figure 6). However, after PSNR value 43, the accuracy decreases more rapidly in scenario 2. Scenario 2's environmental conditions are not as ideal as scenario 1's. Bad weather or dark environments has been shown to reduce the quality of the camera's video *(33, 34)*. Since we use error-bounded lossy compression, we are able to dynamically adapt the loss in the video based on environmental conditions. In the worst case, our scheme reverts to using the raw video footage from the camera. Another alternative to improve detection accuracy is to apply video processing techniques to the video to improve its quality by removing noise or changing the brightness/contrast. We address these issues in our future work. For calculating bandwidth requirements for each PSNR values, we used the following equation:

$$Required\ bandwidth = \frac{S}{T}$$

where $S$ is the size of the video (Mbits) and $T$ is the duration of the video (second).

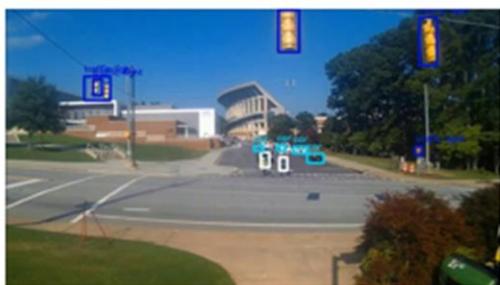
(a) CRF = 10 and PSNR = 56 dB

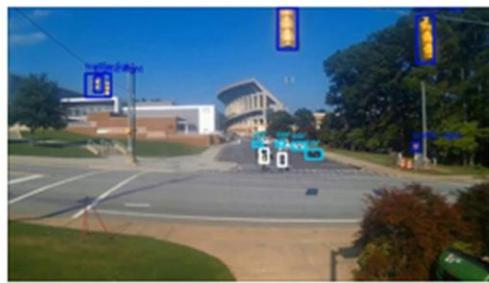
(b) CRF = 20 and PSNR = 49 dB

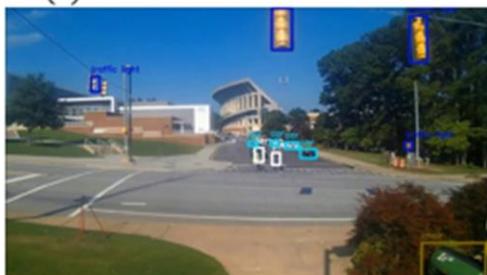
(c) CRF = 30 and PSNR = 43 dB

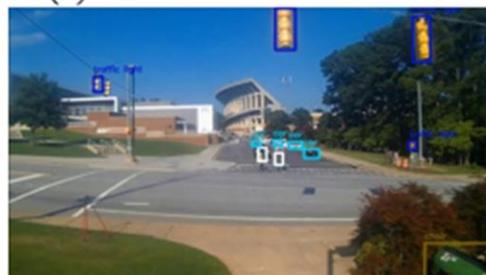
(d) CRF = 33 and PSNR = 41 dB

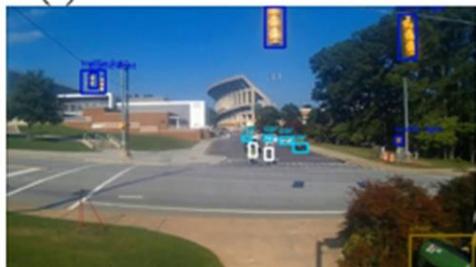
(e) CRF = 35 and PSNR = 40

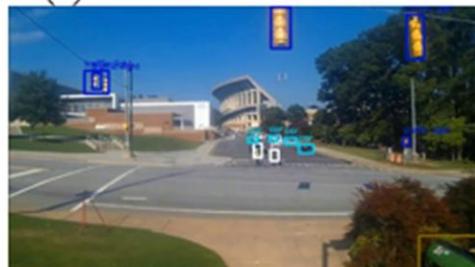
(f) CRF = 37 and PSNR = 39

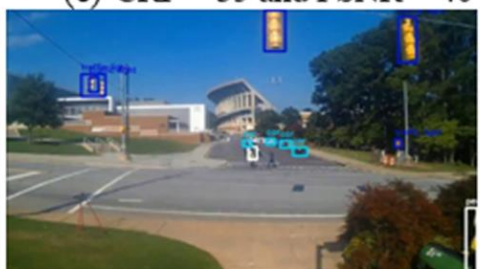
(g) CRF = 40 and PSNR = 37 dB

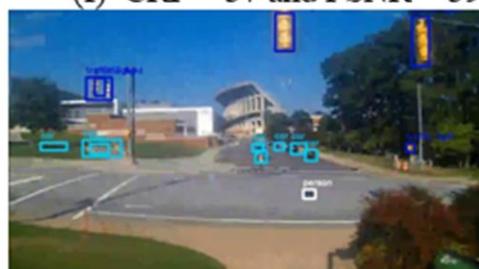
(h) CRF = 50 and PSNR = 31 dB

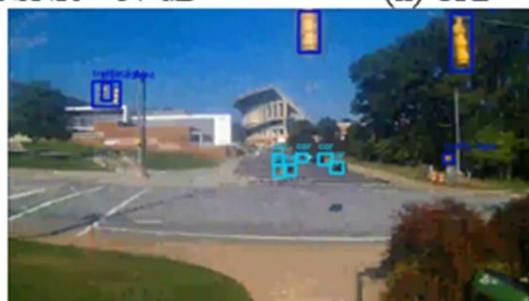
(i) CRF = 51 and PSNR = 30 dB

**FIGURE 5 Pedestrian detection output from YOLO model using increasing levels of video compression and different CRF and PSNR values.**

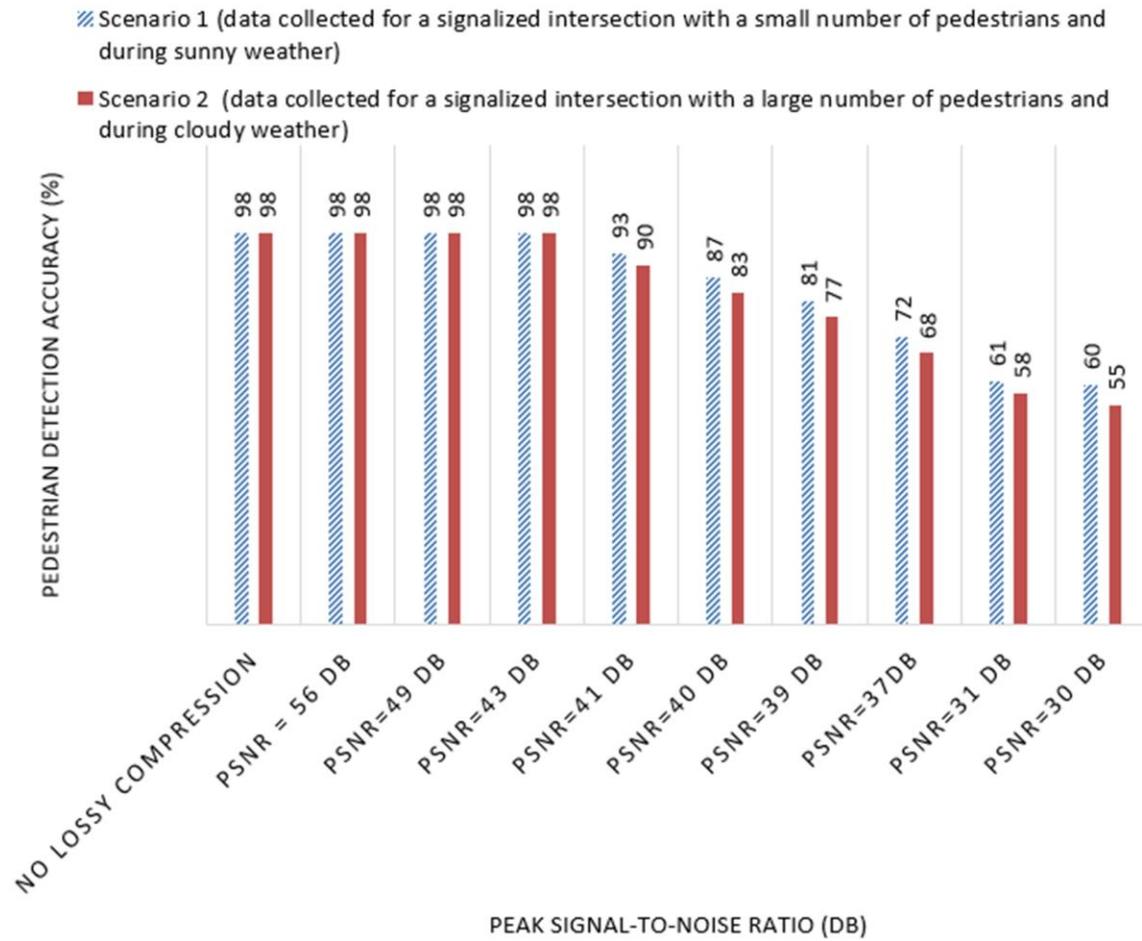

**FIGURE 6 Pedestrian detection accuracy for different PSNR values and scenarios.**

Table 2 shows a summary of the bandwidth requirements for different PSNR values. We calculate the bandwidth requirement for PSNR values that do not cause deterioration in the detection accuracy (56 dB, 49 dB and 43 dB). For these configurations, the corresponding bandwidth requirements are 4.98 Mbits/sec, 1.63 Mbits/sec and 0.31 Mbits/sec, respectively. Comparing to the original bandwidth requirements for the uncompressed video (9.82 Mbits/sec), we are able to achieve the same pedestrian accuracy at only 0.31 Mbits/sec. This results in a 31x reduction in bandwidth required to send the video feed to a processing facility.

**TABLE 2 Communication Bandwidth Requirement for Different PSNR Values**

| CRF value | PSNR Value | Required Bandwidth (Mbits/sec) |
|---|---|---|
| 0 | NA | 9.82 |
| 10 | 56 dB | 4.98 |
| 20 | 49 dB | 1.63 |
| 30 | 43 dB | 0.31 |

Lossy compression technique significantly reduces the storage and bandwidth requirements to archive and transmit the video for different Intelligent Transportation Systems (ITS) applications *(35)*. Reducing storage requirements allows for more minutes of video to be stored or archived with no modification to the underlying hardware *(36)*. Reducing the bandwidth requirement of communicating the video allows for more concurrent video feeds to be sent given a fixed bandwidth; therefore, more cameras can transmit their feeds with no upgrades to the network. Provided the system can compress more frames-per-second (fps) than the video feed creates, the video is compressed in real-time.

In our case study, false positive statistics are available for PSNR value less than 43 dB. As we are not recommending the PSNR value less than 43 dB for compressing video data to maintain a defined pedestrian detection accuracy, there is no impact of false positive for in our approach in the operational environment. However, false positives can be eliminated by using a false positive elimination method, such as object symmetry approach *(37)* for pedestrian detection can be applied to ensure no false positive.

## CONTRIBUTIONS OF THE PAPER

The primary contribution of our paper is to combine a vision-based pedestrian detection model, i.e., the YOLOv3 model, with a lossy compressor for reducing data communication bandwidth while maintaining a defined pedestrian detection accuracy. Our approach highlights the feasibility of using lossy compression to lower the bandwidth requirement for pedestrian detection. Error-bounded lossy compression significantly reduces the storage and bandwidth requirements to archive and transmit the video. Reducing storage requirements allows for more minutes of video to be stored or archived with no modification to the underlying hardware. Reducing the bandwidth requirement of the video feed allows for more concurrent video feeds to be sent given a fixed bandwidth; therefore, more cameras can be transmitting their feeds with no upgrades to the network. Thus, our approach directly contributes to the real-world implementation of pedestrian detection technique with limited bandwidth and storage capacity, as the computation power and storage capacity will be limited at the edge-computing based roadside infrastructure. In addition, using PSNR as our compressed video quality metric allows other researchers to use PSNR values that yield an acceptable combination of detection accuracy and a reduction in the communication bandwidth requirements for evaluating other video compression formats. In addition, an acceptable PSNR values can form a starting point when determining the optimal configuration for integration into a real-world deployment. Future ITS deployments will collect, analyze, and transmit a massive amount of data wirelessly from the CVs, roadside sensors, cell phones, and cameras to roadside edge computing devices in a connected vehicle environment. As the number of connected devices increases, the need for reducing bandwidth requirements grows. This work provides a first step in the exploration and integration of techniques that reduce the bandwidth requirements without sacrificing correctness of the system.

## CONCLUSIONS

Edge computing enables data analytics at the source of the data for real-time safety applications. In this study, we develop and evaluate an edge computing based real-time pedestrian detection strategy combining a pedestrian detection algorithm and an efficient data communication approach to reduce bandwidth while maintaining a high object detection accuracy. We utilize lossy compressed video data at different quality levels to determine the tradeoff between the reduction of the communication bandwidth requirements and a defined object detection accuracy. The performance of the pedestrian-detection strategy is measured in terms of pedestrian classification accuracy with varying PSNRs. The analyses reveal that we detect pedestrians by maintaining a defined detection accuracy (98%) with a peak signal-to-noise ratio (PSNR) of 43 dB while reducing the communication bandwidth from 9.82 Mbits/sec to 0.31 Mbits/sec, 31x reduction in bandwidth. This strategy enables intelligent uses of lossy compression that allows engineers to effectively increase bandwidth and storage capacity enabling them to work with larger quantities of data. Our method is applicable to detect any external objects like vehicles that are not equipped with connected vehicle devices, bicycles, motorcycles, etc. However, pedestrians are the most vulnerable road users. Thus, our research is focused on improving intersection pedestrian safety to show the applicability of our strategy. Future work will evaluate our technique at the other busier intersections with various weather conditions and time of day to show the efficacy of our approach for pedestrian detection. Future work can also consider many unexplored trade-offs important to embedded transportation cyber-physical systems such as energy efficiency, compression/decompression time variation, and effective bandwidth using error bound lossy compression.

**AUTHORS CONTRIBUTION**

The authors confirm contribution to the paper as follows: study conception and design, Mizanur Rahman, Mhafuzul Islam, Mashrur Chowdhury and Jon Calhoun; data collection, Mizanur Rahman and Mhafuzul Islam; interpretation of results, Mizanur Rahman, Mashrur Chowdhury and Jon Calhoun; draft manuscript preparation, Mizanur Rahman, Mashrur Chowdhury, Jon Calhoun and Mhafuzul Islam. All authors reviewed the results and approved the final version of the manuscript.

**ACKNOWLEDGEMENTS**
This material is based on the study supported by the Center for Connected Multimodal Mobility ($C^2M^2$) (USDOT Tier 1 University Transportation Center) Grant headquartered at Clemson University, Clemson, South Carolina, USA. Any opinions, findings, and conclusions or recommendations expressed in this material are those of the author(s) and do not necessarily reflect the views of the Center for Connected Multimodal Mobility ($C^2M^2$), and the U.S. Government assumes no liability for the contents or use thereof.